\documentclass[acmsmall]{acmart}

\usepackage{verbatim}
\usepackage[]{algorithm2e}
\usepackage{graphicx}
\usepackage{booktabs}
\usepackage{relsize}
\usepackage{colortbl}
\usepackage{enumerate}
\usepackage{amsmath}
\usepackage{tikz}
\usepackage{pgf}
\usepackage{pbox}
\usepackage{rotating}
\usepackage{multirow}
\usepackage{balance}
\usepackage{subfigure}
\usepackage{listings}
\usepackage{bbding}
\usepackage{fancybox}
\usepackage{url}
\usepackage{pifont}
\usepackage{threeparttable}
\usepackage{soul}
\usepackage{adjustbox}
\usepackage{cleveref}[2012/02/15]
\usepackage{arydshln}
\crefformat{footnote}{#2\footnotemark[#1]#3}
\usepackage{xcolor}
\usepackage{makecell}
\usepackage{threeparttable}
\usepackage[T1]{fontenc}

\usepackage{soul}

\renewcommand\refname{REFERENCES}

\renewcommand{\labelitemi}{--}
\newcommand{\startlist}{\begin{list}{\labelitemi}{\leftmargin=1em}\setlength{\itemsep}{-1mm}}
\newcommand{\stoplist}{\end{list}}

\newcommand{\smallsection}[1]{\noindent {\bf \underline{#1}}.\hspace{1mm}}

\newcommand{\hypobox}[1]{\begin{center}%
	\noindent\thicklines\setlength{\fboxsep}{7pt}%
	\cornersize{0}\Ovalbox{\begin{minipage}{0.9\columnwidth}%
	\vspace{-0.1cm}
	\textit{#1}
	\vspace{-0.1cm}
	\end{minipage}} \end{center}}

\newcommand{\compnany}[1]{{{ [Company X]}}}

\setcopyright{acmcopyright}

\setcopyright{acmcopyright}

\acmJournal{TOSEM}
\acmVolume{}
\acmNumber{}
\acmArticle{}
\acmMonth{8}

\begin{document}

\title{Can I use this publicly available dataset to build commercial AI software?-A Case Study on Publicly Available Image Datasets }


\author{Gopi Krishnan Rajbahadur}
\email{gopi.krishnan.rajbahadur1@huawei.com}
\affiliation{%
  \institution{Centre for Software Excellence, Huawei Canada}
  \state{ON}
  \country{Canada}}

\author{Erika Tuck}
\email{etuck@yorku.ca}
\affiliation{%
  \institution{Lassonde School of Engineering, York University}
  \state{ON}
  \country{Canada}}
  
 \author{Li Zi}
\email{lizi4@huawei.com}
\affiliation{%
  \institution{Huawei China}
  \state{ON}
  \country{China}} 

\author{Dayi Lin}
\email{dayi.lin@huawei.com}
\affiliation{%
  \institution{Centre for Software Excellence, Huawei Canada}
  \state{ON}
  \country{Canada}}

\author{Boyuan Chen}
\email{boyuan.chen1@huawei.com}
\affiliation{%
  \institution{Centre for Software Excellence, Huawei Canada}
  \state{ON}
  \country{Canada}}

\author{Zhen Ming (Jack) Jiang}
\email{zmjiang@cse.yorku.ca}
\affiliation{%
  \institution{Lassonde School of Engineering, York University}
  \state{ON}
  \country{Canada}}

\author{Daniel~M.~German}
\email{dmg@uvic.ca}
\affiliation{%
  \institution{University of Victoria}
  \state{ON}
  \country{Canada}}
\renewcommand{\shortauthors}{Rajbahadur et al.}


\begin{abstract}

Publicly available datasets are one of the key drivers for commercial AI software. The use of publicly available datasets is governed by dataset licenses. These dataset licenses outline the rights one is entitled to on a given dataset and the obligations that one must fulfil to enjoy such rights without any license compliance violations. Unlike standardized Open Source Software (OSS) licenses, existing dataset licenses are defined in an ad-hoc manner and do not clearly outline the rights and obligations associated with their usage. Further, a public dataset may be hosted in multiple locations and created from multiple data sources each of which may have different licenses. Hence, existing approaches on checking OSS license compliance cannot be used. In this paper, we propose a new approach to assessing the potential license compliance violations if a given publicly available dataset were to be used for building commercial AI software. We conduct a case study with our approach on 6 commonly used publicly available image datasets. Our results show that there exists potential risks of license violations associated with all the studied datasets if they were used for commercial purposes. 


\end{abstract}



\maketitle

\section{Introduction}
\label{sec:introduction}

The adoption and the commercialization of Artificial Intelligence (AI) in software has increased significantly in the past decade~\cite{gartner2019}. One of the key components of AI software is AI models, which are trained from large swathes of data. Companies looking to build commercial AI software can acquire datasets through several approaches such as creating and curating data from scratch~\cite{drexl2016data}; acquiring data from external sources such as research publications in the form of publicly available datasets~\cite{benjamin2019towards}; crawling and curating datasets from publicly available data sources; and, purchasing datasets through dataset vendors~\cite{agarwal2019marketplace}. Among these approaches, one of the most popular approaches is to use publicly available datasets to build commercial AI software~\cite{GluonCVYOLOv3,ModzyModel,TensorFlowdeeplabv3,peng2021mitigating}. 



It is important to ensure that the usage of publicly available datasets, particularly for commercial purposes, remains legally compliant~\cite{benjamin2019towards}. The usage of a publicly available dataset is governed by the dataset license associated with the dataset (much like Open Source Software (OSS)). Similar to OSS licenses, dataset licenses outline the \emph{rights} that users are entitled to and the \emph{obligations} that users have to fulfill in order to enjoy these rights. From an AI software development perspective, the rights outlined by a particular dataset license determines whether the dataset can be used in a specific usage context (e.g., for building commercial applications or to redistribute). The obligations under a dataset license can be considered as software requirements (e.g., redistributing the AI software under the same license) that need to be captured and traced throughout the development lifecycle of the AI software, if the dataset is used. Therefore, similar to the usage of OSS in commercial software, to ensure legally compliant commercial usage of a dataset, it is important to adhere to the rights and obligations outlined in the dataset's licence (i.e., dataset license compliance). 

However ensuring dataset license compliance poses several unique challenges in contrast to OSS license compliance. First, it is difficult to locate and identify the complete and correct license(s) associated with a given dataset. For instance, as Peng et al.~\cite{peng2021mitigating} state, many retracted datasets continue to be available due to lack of clarity in retraction. Such a case makes it difficult to identify and locate the latest license of a dataset. Second, verifying the validity of the license associated with a given dataset is challenging. Many publicly available datasets are created by combining data from multiple data sources (e.g., websites or other datasets). Each of these data sources may have different licenses (e.g., website's terms of use). However, the creators of publicly available datasets seldom document the different licenses associated with the different data sources that they use. In addition, they do not consider the impact of these different licenses on the license that they assign to the aggregated dataset (which might make the license potentially invalid or wrong). For instance, CIFAR-10 dataset's license grants all rights to users and the only obligation that the license outlines is a requirement for citation. However, CIFAR-10 is created by crawling images from different sources such as Google Images and Flickr which may have licenses that restrict the usage of the images in any commercial context. Therefore, considering only CIFAR-10's license to ensure compliance in such a case might be problematic. Finally, as prior studies note~\cite{benjamin2019towards,peng2021mitigating}, the licenses of publicly available datasets are typically ambiguous and do not clearly outline the rights and obligations that govern the usage of datasets. Such lack of clarity makes it difficult in practice to use the datasets to build commercial AI software without the risk of license violation. For instance, GitHub Copilot~\cite{GitHubCopilot} is an AI assistant that suggests whole lines of code (or sometimes whole functions). It leverages a large AI model that was trained on billions of lines of source code that are hosted on GitHub. While usage of open source code as a computer program is well-governed by OSS licenses and are common among GitHub users, the right to use code as data, in particular in the context of training AI models for commercial purposes, is not clearly defined in the OSS licenses. Such ambiguity in licenses has led to wide legal debate on the compliance of GitHub Copilot~\cite{TheVergeGitHubCopilot, JonesGitHubCopilot, RedaGitHubCopilot}. 

In this paper, we propose an approach to extract and analyze the compliance of dataset licenses. Our approach can help AI engineers extract the rights, 
and the ensuing obligations under a license. 
AI engineers can leverage our analysis results to evaluate any potential risks associated with using such a dataset under their specific usage context. We showcase the usefulness of our approach through a case study on 6 widely used publicly available image datasets (CIFAR-10, ImageNet, Cityscapes, MS COCO, FFHQ and VGGFace2). We found that although these datasets were used widely in many of prior research~\cite{fan2021makes} and commercial endeavors~\cite{GluonCVYOLOv3,ModzyModel,TensorFlowdeeplabv3,peng2021mitigating}, all of our studied datasets might not be suitable to build commercial AI software due to a high risk of potential license violations. Our findings highlight that AI engineers need to be aware of potential license-related risks and problems to avoid legal issues. The main contributions of our paper are: 

\begin{enumerate}
   \item This is the first work that proposes an approach to assisting practitioners on assessing the risks of using publicly available datasets to build commercial AI software. 
   
    \item As part of our approach, we have proposed several data schema which serves as a standardized way for documenting various dataset license related information that was collected during our analysis. These schema facilitate easy sharing of the analysis results of various datasets (e.g., different AI usage contexts or interpretation of different dataset licenses) and streamline the adoption of our approach. We have provided a replication package with the schema and results of our analysis documented with these schema to facilitate the sharing and the verification of our results\footnote{https://github.com/TheTuckStopsHere/Replication-package}.
   
   \item Through our case study, we present the potential license compliance violations that one may run into when using six  publicly available datasets in a commercial context. 
   Furthermore, based on our experience on analyzing license compliance for various publicly available datasets, 
   we also provide recommendations for AI engineers who use publicly available datasets to build AI software.
   

\end{enumerate}

The rest of the paper is organized as follows. Section~\ref{sec:background} presents the relevant background and related work of our study. Section~\ref{sec:approach} presents the approach that we use to identify potential license compliance violations that may occur when using publicly available datasets to build AI software. Section~\ref{sec:casestudy} presents the setup and results of our industrial case study. Section~\ref{sec:feedback} provides the feedback from the industry practitioners. Section~\ref{sec:recommendation} provides several recommendations for AI engineers in light of our findings. Section~\ref{sec:threats} documents threats to validity and Section~\ref{sec:conclusion} concludes our study. 
\section{Background and Related Work}
\label{sec:background}

In this section, we describe the background details and prior work that is related to our study.

\subsection{Legal rights and protections on datasets}
In this sub-section, we highlight some of the legal protections that govern and protect the usage of datasets in a commercial context. A dataset and the data points contained in the dataset are typically governed by two important laws: copyright law and contract law. Though the actual terms of the protection offered by these laws vary across countries, these laws in general provide several protections that we details below. In addition, with the recent advent of AI, there have been several privacy and fairness related laws that also govern the usage of a dataset, which we detail later in this sub-section. 

\smallsection{Copyright law} Copyright law prevents anyone from copying or reproducing any part of a work (e.g., an image, video) without explicit permission from the owner of the copyright~\cite{canadacopyright}. Therefore, by default, copyright-protected data cannot be used commercially or distributed unless explicitly allowed by the copyright holder. However, many publicly available datasets are known to contain copyrighted data~\cite{benjamin2019towards,peng2021mitigating}. Using them to build commercial AI software could potentially result in copyright infringement. 

However, it is important to note that in certain cases and countries it might be acceptable to use copyright-protected data for various purposes including commercial purposes without explicit permission from the copyright holders. For instance, in the United States, as the recent lawsuit \textit{Authors Guild v. Google}~\cite{lawsuit} suggests, when there is no material damage done to the copyright holders, under the Fair Use doctrine~\cite{USFairUse} it is acceptable to use copyrighted data for commercial purposes, even without explicit permission from the copyright holders. However, in other countries the same may not be true. In the United Kingdom and Canada, for instance, copyright-protected data can only be used for non-commercial purposes without explicit permission from the copyright holders under the fair dealing exception to copyright infringement~\cite{CanadaFairDealing}. Similarly, in the EU, under the Text and Data Mining Law~\cite{triaille2014study}, copyright-protected material can be used without explicit permission from the copyright holders only for non-commercial purposes. In other parts of the world, different copyright laws may restrict the usage of copyright-protected materials in a commercial context. In summary, though there are certain exceptions that allow one to use copyright-protected data in a commercial context, it usually varies on a case-by-case basis. Therefore, using publicly available datasets with copyright-protected data to build commercial AI software could result in potential copyright infringement.
 
\smallsection{Contract law}
Contract law states that owner of the copyright of a work (e.g., images, videos) may grant a license which outlines the rights that others can enjoy and the obligations that they must fulfil to enjoy those rights. When the terms of the license are not respected, i.e., if a right that is not granted by the license is exercised on the data or if the obligations are not fulfilled, then the potential violation may amount to breach of contract or contract violation. Depending on the country where the dataset was created or is intended to be used, either copyright law or contract law takes precedence.
 
 
In addition to the aforementioned protections, many countries around the world have regulations that dictate how private information in datasets can be fairly leveraged~\cite{EuropeanCommissionEthics, PrivacyCommissionerSubmission, EuropeanCommissionProposal}. Though all privacy, fairness, copyright and license related issues can cause compliance issues, in this paper we only focus on identifying the potential license compliance related issues that arise from using publicly available datasets. The potential risks that we assess does not necessarily constitute as legal risks. We simply propose an approach to identify the potential risks. 
 


\subsection{Objectives and challenges associated with analyzing dataset license compliance}
The key objective for a company to analyze the license of a publicly available dataset is to understand and determine if the given dataset can be used in a specific usage context (e.g., model training, redistribution) commercially while ensuring license compliance. License compliance is a form of regulatory compliance. Ensuring regulatory compliance in AI software is pivotal across the globe~\cite{ingolfo2013arguing,breaux2008analyzing,kiyavitskaya2008automating,zeni2015gaiust}. When an AI software uses a publicly available dataset, it is implicit that the owners of the AI software enter into an agreement with the copyright holders of the dataset. Such an agreement entails that the AI software uses the dataset only per the rights provided by the license and fulfils the obligations outlined in the license, which makes the obligations a software requirement~\cite{ingolfo2013arguing}. If the AI software fails to do so, it will be in breach of the dataset's license, which becomes an issue of regulatory compliance. As several prior studies note, organizations that are found non-compliant of laws and regulations face serious consequences~\cite{ingolfo2013arguing,breaux2008analyzing,kiyavitskaya2008automating,zeni2015gaiust}. Therefore, compliance is both a functional and non-functional requirement of an AI software~\cite{breaux2008analyzing}. 

However, as Benjamin~et~al.~\cite{benjamin2019towards} outline, licenses associated with publicly available datasets do not clearly state the rights and obligations associated with the usage of datasets. Such a problem poses a challenge to AI engineers trying to encode obligations as requirements. Ensuring license compliance under such ambiguous licenses is challenging as we explain in Section~\ref{sec:introduction}. In cases such as those as Breaux~et~al.~\cite{breaux2008analyzing} point out, when requirements, in particular legally enforceable requirements, are ambiguous, it is important for AI engineers to do their \textit{due diligence} or take \textit{due care} to make sure their systems are compliant. In other words, AI engineers should make reasonable efforts to make sure that AI software is not found in breach of licenses. To make such a claim, AI engineers should employ traceability from the rights and obligations in the license to software requirements~\cite{breaux2008analyzing,kiyavitskaya2008automating}. In addition, it is imperative for AI engineers to justify that their interpretation of the rights and obligations is consistent with how they implemented those requirements in the constructed AI software~\cite{breaux2008analyzing}. Therefore, a rigorous approach to ensure license compliance, like the one we propose in our paper that ascertains the rights and obligations associated with a publicly available dataset, is vital. 

\vspace{-0.5cm}
\subsection{Related work}

\smallsection{Assessing license compliance violations} Though there is no prior work that studies the license compliance violations that happen when publicly available datasets are used to build commercial AI software, there is a rich body of work that studies license compliance violations when reusing OSS. Several prior works focus on how to assess license inconsistencies (i.e., violating the license of the OSS component that is being reused) that occur in OSS~\cite{wu2017analysis,wu2015method}. For instance, several studies focus on designing automated tools to assess licence compliance~\cite{zhang2010automatic,van2014tracing,german2012method,german2009license,kapitsaki2017automating}. Despite there being several techniques to ensure OSS license compliance, they cannot be directly used to assess license compliance violations that happen when using publicly available datasets. As discussed previously, unlike OSS, publicly available datasets have unclear licenses and license validity issues. Furthermore, as German and Di Penta~\cite{german2012method} show, to check license compliance it is important to identify a software component's provenance and lineage (and they outline methods to do so). However, identifying a dataset's provenance and lineage is not as straightforward. Therefore, an approach to assessing dataset license compliance violations will be quite different from that of OSS. This is the main focus of our paper. 


\smallsection{Software regulatory compliance} Similar to complying with the terms of a license, software also needs to comply with certain regulatory requirements. For instance, software built to handle information about patients in the United States needs to comply with HIPAA~\cite{HIPAA}. The rules outlined by such regulations are requirements (similar to rules outlined by a license) that must be followed by a software to which the given regulation is applicable~\cite{breaux2008analyzing}. There are several studies that focus on how to ensure that the requirements outlined by a regulation are followed by the software under consideration~\cite{breaux2008analyzing,breaux2008legal,kiyavitskaya2008automating,zeni2015gaiust}. The first step in checking if the software complies with the regulation is to identify the rights and obligations outlined by the regulation. Zeni~et~al.~\cite{zeni2015gaiust} propose a tool called GauisT to semi-automatically identify the rights and obligations of a regulation. Breaux~et~al.~\cite{breaux2008analyzing} provide an alternate method to extract the rights and obligations from a regulation to enable checking if the software is indeed compliant with the regulation. They showcase the usefulness of their approach by checking if a software is compliant with HIPAA. Unfortunately, we cannot use the methods outlined in these studies as these methods presuppose that a regulation clearly outlines the rights and obligations and that the regulations are valid. However, in our case, the licenses associated with publicly available datasets are typically unclear and have concerns regarding their validity. 

\smallsection{Licenses for publicly available datasets} There are a few dataset specific licenses that are available, and these licenses can largely be grouped into \emph{copyleft} and \emph{permissive} licenses. The copyleft dataset licenses typically require the AI software using the dataset to be licensed under the same license under which the dataset was provided (or they do not clearly mention what license the software built using these datasets should have). Some examples of copyleft licenses are Community Data License Agreement (CLDA) sharing 1.0 and Creative commons (CC)-By-SA (Share Alike), CC-BY-NC-SA. In contrast, permissive license allow the AI software that uses these datasets to be licensed under any license one wishes. Example of permissive licenses are CC-BY (public domain), CDLA permissive. However, as Benjamin~et~al.~\cite{benjamin2019towards} note even these licenses are not standardized and their rights and obligations are typically ambiguous.  In addition to showcasing the shortcomings of standard dataset licenses, they provide a new license called the Montreal Data License (MDL), which is more suitable for datasets. Different from their study~\cite{benjamin2019towards}, we focus on assessing potential license compliance violations if certain datasets are used in a commercial context.

\section{Approach}~\label{sec:approach}

\vspace{-0.7cm}
\begin{figure}[htp]
    \caption{An overview of our approach.}
    \centering
    \includegraphics[width=0.6\textwidth]{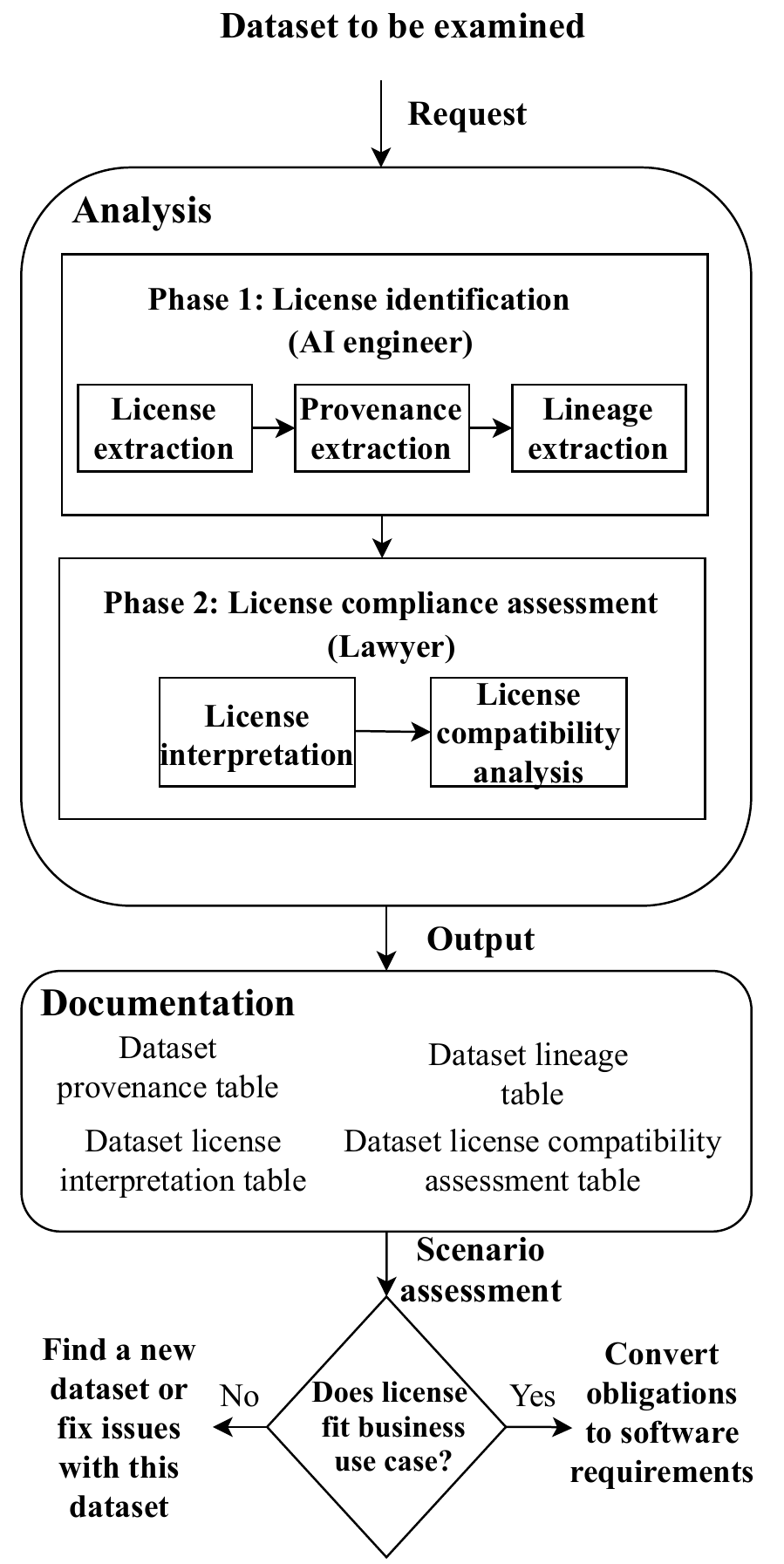}
    \label{fig:overall}
  
\vspace{-0.5cm}
\end{figure}

In this section, we describe our approach for identifying whether using a publicly available dataset to build a commercial AI software would result in a license compliance violation. It starts with an AI Engineer initiating the request for license compliance analysis for a particular dataset. The license compliance analysis approach consists of two phases: Phase 1, called \textit{License identification}, involves gathering the information we need about a dataset’s license, provenance and lineage. 
Phase 2, called \textit{License compliance assessment}, involves interpreting the output of Phase 1 and then completing a risk assessment based on the findings.
Phase 1 can be done by an AI engineer; however, Phase 2 must be done by a lawyer, as interpreting the information contained in the collected licensing information requires legal expertise~\cite{gordon2014role,german2009license,german2012method}. Over the course of our approach, various information (e.g., a dataset's provenance, lineage) are documented in a database. Once the analysis is complete, the AI Engineer uses final dataset license compatibility assessment table (the final table that is produced from our approach) to ascertain if the given dataset can be used commercially for the usage scenario they are interested in. 

It is important to note that when an AI Engineer initiates a request to analyze the license compliance risks associated with a dataset, we first check if we already have conducted the analysis by looking up the results from the previous analyses. Only if we haven't analyzed the given dataset before, will we go through our proposed approach. Else, we simply return the stored license compatibility assessment table from the last time we assessed the given dataset.

In the remainder of this Section, we explain both phases and the steps and sub-steps contained in each phase. As a running example, we analyze the license compliance of CIFAR-10 dataset~\cite{CIFAR, KagglePopDatsets}.


\subsection{Phase 1: License Identification}
In this phase, an AI engineer identifies all the license details that need to be interpreted for potential license compliance violations. For the running example, the License Identification phase was individually carried out by the first three authors of the study to ensure correctness. The results of all the steps were considered final only after all three authors agree.

\smallsection{(Step 1) License extraction} Unlike OSS source code, dataset licenses are not made available in a standard fashion. Therefore, we first search for the license on the website from which the dataset was downloaded. If we are unable to find a license, we then check if a license is provided as a separate file as part of the dataset. If not, we reach out to the owners of the dataset asking about the dataset's license. 

\noindent\textsf{Running example.} In this paper, we download the CIFAR-10 dataset from its official website. Also on the CIFAR-10 website, we find the following request from the dataset creators: ``Please cite~\cite{krizhevsky2009learning} if you intend to use this dataset," alongside a link to the paper. We extract this as the dataset's license. 

\smallsection{(Step 2) Provenance extraction}
The provenance of a dataset represents the true source of origin of the dataset~\cite{godfrey2015understanding}. A dataset could be first created by some researchers and could later be distributed through different platforms (sometimes even with a different license than the one assigned by the creators). In addition, it is important to verify if the dataset we obtained is indeed the same dataset created by the creators of the dataset. Therefore, it is pivotal to ascertain the provenance of a dataset to verify if the license that we extracted in the previous step is indeed the correct license of the dataset. 

\noindent\emph{{(Sub-step 1)} Locate official dataset source.} The first task is to locate the official source of the dataset (e.g., official website, research paper, or technical report) by querying search engines with appropriate search terms. 

\noindent\textsf{Running example.} We have already located the official source of CIFAR-10, which we verify through the presence of the accompanying publication.

\noindent\emph{{(Sub-step 2)} Extract license and metadata of official source.}    
We then extract the license and certain metadata about the dataset by collecting the details that we outline in the provenance schema, including the contents of the license. (See Table~\ref{tab:provenance}, which shows the recorded provenance of CIFAR-10.). For extracting the license, we recommend official websites as the first place to look for the license. Sometimes this requires creating an account. If a license is not found on the official website, then one can download the dataset to see whether the license is packaged with the dataset files.
Once we extract a dataset’s provenance, if the license that we extracted in the previous step contradicts the license extracted in this step, we use the license that we extracted through provenance as the dataset’s license.

\noindent\textsf{Running example.} 
While extracting CIFAR-10's provenance, we learn that the authors of the website that hosts the dataset~\cite{CIFAR} and the authors of the associated publication are the same. As such, we simply collect the dataset's provenance details as shown in Table~\ref{tab:provenance}. Note that detailed explanation of our schema can be found in our replication package. In addition, since the license extracted in this and the previous step are consistent, we know the license that we extracted in the previous step is the dataset’s license.

\smallsection{(Step 3) Lineage extraction}
A dataset’s lineage tracks the data sources from which a given dataset was created~\cite{allen2015multi}. Many publicly available datasets, including computer vision and NLP datasets, are created by collecting data from different sources, which are generally other datasets or popular websites that host or index data points such as images. Not surprisingly, these different data sources typically have a license that is different from that of the dataset but are no less important to verifying the validity of the license of the dataset under examination. It is important to note that if a dataset is completely created from scratch (i.e., not extracted from any data source), then we do not have any lineage details for that dataset. 

\noindent\emph{(Sub-step 1) Trace dataset creation process.} We want to know how a dataset was created, which requires that we trace all of its data sources. The information we need to begin this sub-step is stored in the “Description of the data collection process” field that we recorded when we extracted the dataset’s provenance. Then, if we learn that this data source in turn has other data sources, we also record those data sources recursively.

\begin{figure}[htp]
    \caption{Recorded lineage details for CIFAR-10 and its data sources (for each data source, we collect the provenance details that we mention in Table~\ref{tab:provenance})}
    \centering
    \includegraphics[width=0.6\textwidth]{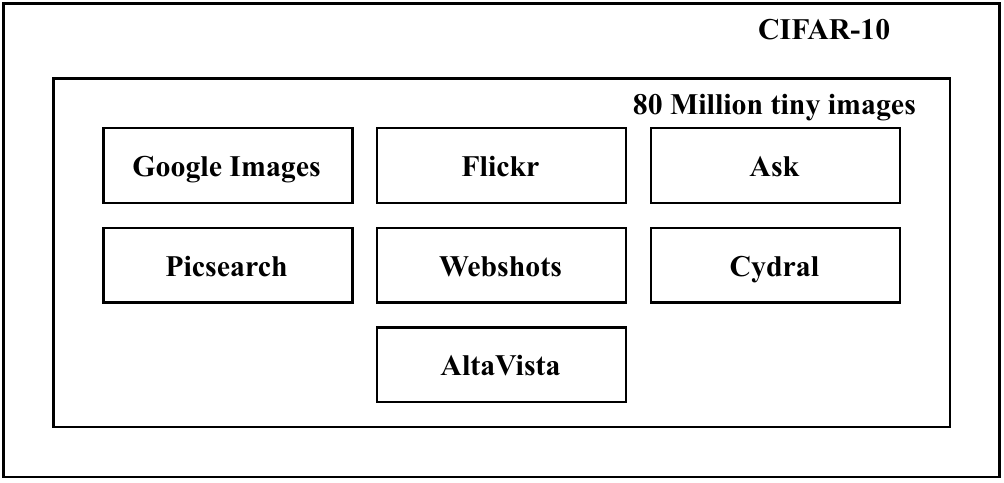}
    \label{fig:lineage}
\end{figure}

\noindent\textsf{Running example.} 
We previously learned through provenance extraction that CIFAR-10 is a subset of another dataset called 80 Million Tiny Images. Reading the research paper associated with 80 Million Tiny Images~\cite{torralba200880}, we learn that this dataset has 7 data sources: Google, Flickr, Ask, Altavista, Picsearch, Webshots and Cydral. Figure~\ref{fig:lineage} is a visual representation of CIFAR-10’s data sources.

\noindent\emph{(Sub-step 2) Locate official sources of data sources.} We verify information about the data sources by locating their official sources (e.g., official dataset website, search engine and so on) similar to sub-step 1 of Provenance extraction.

\noindent\textsf{Running example.} First, we locate the official dataset website for 80 Million Tiny Images~\cite{80M} through Google search. After learning that the dataset has been withdrawn, we then locate the archived versions of the website available on the Internet Archive~\cite{80MWebArchive}. Following that, we visit the official websites for each of the 7 data sources listed above.

\noindent\emph{(Sub-step 3) Determine the license range.} 
A key challenge in determining the correct license of a data source is that a dataset might have been created some time ago. Since then, the license of the data source might have evolved. To mitigate this problem, we identify the license range associated with a given dataset. License range refers to the time period when the data contained in the dataset was likely collected. This, of course, is merely an estimate, but it is important because if a dataset’s contents come from a data source (e.g., Google Images), then we need to know which past versions of the data source’s license are relevant to the dataset (e.g., Google’s Terms of Service from 2005, and/or 2007, and/or 2012, and so on). Only the licenses that were effective during the license range of the dataset are of interest to us. 

To determine the end of the license range, we use the \textit{Origin date} field that we collect during the provenance extraction stage, which identifies the year when the dataset first originated. A variety of chronological information about a dataset can be used as a proxy for the origin date, including when the dataset’s research paper was first published, when the dataset was first made available as part of a programming challenge, and so on. We consider the year preceding the origin date to be the beginning of the license range. Data sources such as websites and search engines inherit their license ranges from the dataset to which they contribute content. However, the license range for data sources that are datasets themselves is based on the dataset's origin date.

\noindent\textsf{Running example.} During provenance extraction, we learn that publication year for CIFAR-10 is 2009~\cite{krizhevsky2009learning}, which we determine to be the \textit{Origin date}. So we set the license range to be 2008 - 2009. Its top-level data source, 80 Million Tiny Images, was created in 2006~\cite{80M}. So its license range is 2005 - 2006. Then, the 7 data sources of 80 Million Tiny Images inherit the same licensing range as their parent dataset, 80 Million Tiny Images.

\noindent\emph{(Sub-step 4) Identify license for data sources.} Using a Web archive~\cite{WebArchive}, we collect the earliest available license of each data source during the license range. We do this recursively until we identify the license associated with all the data sources that contributed to the creation of the dataset under consideration. Along with the licenses, for each data source, we collect the provenance details that we outline in Table~\ref{tab:provenance}. We use a web archive because, the license as it appeared when the dataset was first released is the one that applies.

\noindent\textsf{Running example.} No license captures were available during the license range for CIFAR-10 and for two of its data sources (Ask and Webshots), so we simply store the earliest captures we can find. The license content of 80 Million Tiny Images and Cydral is completely unavailable, and so, we store \textit{N/A}. For the remaining data sources, we record the earliest license captures from the correct license range and its provenance details. We do not show the provenance details recorded for each data source due to space constraints.

\begin{table*}[htbp]
  \centering
\scriptsize
 
\caption{Recorded provenance details for CIFAR-10} ~\label{tab:provenance}

\begin{tabular}{|p{1cm}|p{2cm}p{2cm}p{2cm}p{4cm}|}
\hline
\multirow{2}{*}{\textbf{\makecell[l]{Dataset-\\related\\ details}}} & \textbf{Dataset name}  & \textbf{Dataset version} & \textbf{Origin date} & \textbf{Origin}\\
\cline{2-5}
& CIFAR-10 & N/A & 2009 & \url{https://www.cs.toronto.edu/~kriz/cifar.html}\\
\cline{2-5}
 & \multicolumn{2}{l}{\textbf{Description of dataset}} & \multicolumn{2}{l|}{\textbf{Description of data collection process}}   \\

\cline{2-5}
 & \multicolumn{2}{l}{\makecell[l]{The CIFAR-10 dataset consists of 60000 32x32 \\colour images in 10 classes, with 6000 images per class.\\ There are 50000 training images and 10000 test images}} & \multicolumn{2}{l|}{\makecell[l]{The CIFAR-10 and CIFAR-100 are labeled subsets\\ of the 80 million tiny images dataset. They were collected by \\Alex Krizhevsky, Vinod Nair, and Geoffrey Hinton.}}\\

\cline{2-5}
& \textbf{Downloaded outlet} & \textbf{Is outlet licensed?} & \textbf{Is dataset publicly available?} & \textbf{Additional notes} \\
 \cline{2-5}

& N/A & N/A & Yes & \makecell[l]{This dataset is a subset of another dataset\\ called 80 Million Tiny Images}\\
 \cline{2-5}

\hline
\multirow{2}{*}{\textbf{\makecell[l]{License-\\related\\ details}}} & \multicolumn{2}{l}{\textbf{Where license was found}} & \textbf{License location} & \textbf{License content} \\
\cline{2-5}
& \multicolumn{2}{l}{Present on the official dataset website} & \url{ https://www.cs.toronto.edu/~kriz/cifar.html} & (not pasting content due to space) \\
\hline 
\multirow{2}{*}{\textbf{Metadata}} & \multicolumn{2}{l}{\textbf{Hashcode}} & \textbf{Size} & \textbf{Format} \\
\cline{2-5}
& \multicolumn{2}{l}{MD5: c58f30108f718f92721af3b95e74349a (Python version)} & 163MB (Python version) & tar.gz \\

\hline
\end{tabular}
\end{table*}

\subsection{Phase 2: License compliance assessment.} Once an AI engineer identifies the license associated with a dataset and the licenses of the data sources that were used to create the dataset, a lawyer decomposes each license to identify the rights and obligations it outlines. As several prior studies in license and regulatory compliance~\cite{gordon2014role,german2009license,german2012method} illustrate, software and legal expertise is required for this type of task. Therefore, in our paper we have consulted a lawyer to help us interpret these extracted licenses and complete Phase 2. 


\smallsection{(Step 1) License interpretation} The first step of Phase 2  requires a lawyer to read the license associated with the dataset and the licenses of the data sources from which the data were collected. For each license, the lawyer then extracts the rights and obligations outlined in it. 

In order to ensure that the extracted rights and obligations are documented in a standard format that facilitates understanding, we document them using a modified version of the Montreal Data License (MDL) proposed by Benjamin~et~al.~\cite{benjamin2019towards}. The reason is because MDL standardizes and documents what rights are granted to the dataset itself (e.g., modify, distribute) and what rights are granted to the dataset in conjunction with AI software (e.g., train, research). These rights can be directly linked to the usage scenarios. For example, if the dataset license grants rights to distribute the dataset, then the usage scenario of distributing the dataset will be permissible. The MDL was originally proposed to create licenses for sharing new datasets. However, for this work, we re-purpose the MDL format to document the legal interpretations associated with datasets (i.e., to decompose licenses into their rights and obligations) and identify the usage scenarios that are acceptable. We further enhance MDL in the following three aspects: 
\begin{enumerate}
\item The MDL does not record the obligations that the user of a dataset needs to fulfil when enjoying a specific right. 
Therefore, we enhance the MDL by capturing both rights and obligations for all fields in \texttt{Rights to dataset} and \texttt{Rights to dataset in conjunction with AI models}.
\item When recording \texttt{Rights to dataset in conjunction with AI models}, the MDL does not capture if the license allows reverse engineering the trained AI model to reconstruct the data. As such, we add a new field ``Model reverse engineer'' to record this. 
\item We add fields to record license validity period and the liability or warranty provided by the license. 
\end{enumerate}
We call our modified MDL documentation schema the \texttt{Enhanced MDL}. We provide the schema and a description of each field in Table~\ref{tab:MDL} in our replication package. 

\noindent\textsf{Running example.} The main results of our decomposition can be seen in Table~\ref{tab:MDL}. The fields marked in blue represent our enhancement. In summary, a user has rights to access the CIFAR-10 data for every use case outlined in the schema. To do so, the user is required to cite the associated technical report, as requested on the official dataset website.

\noindent\emph{(Optional sub-step) Documentation and schema update.} In the event that one needs to ascertain if a new usage scenario other than the ones that are dictated by the rights that our enhanced MDL schema captures, then they should enhance our schema. They can do so by adding a column to capture the rights and obligations that would pertain to their usage scenario. 

\noindent\textsf{Running example.} In case we wanted to understand if we could build adversarial models with the given dataset, we would enhance our schema to include rights and obligations required to use the dataset to build an adversarial model. 

\smallsection{(Step 2) License compatibility analysis} Based on the well formatted information of the Enhanced MDL, a risk assessment is performed to decide whether the dataset should be used in the commercial context for which it is being considered. The risk assessment, which involves interpreting the  compatibility between multiple legal licenses, advises practitioners on what rights they can enjoy commercially and what obligations they have to satisfy to do so. These rights in turn inform the practitioners what usage scenarios they can use the dataset for, and what are the obligations that they need to fulfil to do so.

Before assessing the usage scenarios that are permitted, we need to ascertain the final rights and obligations that are associated with the given dataset after taking the licenses of the data sources into consideration. To reconcile the rights and obligations of using a dataset, we need to match use cases across the different interpreted licenses we have collected. We employ the following heuristics: 
\begin{enumerate}
\item If a dataset license and all its data source licenses agree that a use case is allowed, then the action is permitted. This permission is reflected in the verified license.
\item If the license of any data source restricts a use case then, even if the main dataset license allows it, the verified license records this discrepancy by marking the action as not permitted.
\end{enumerate}

\noindent Essentially, we aim to verify if the rights and obligations of the data sources allow for the dataset license to be valid. For instance, a very permissive dataset license cannot be valid if one of the data sources has a more restrictive license, since the license of the dataset would no longer respect the license under which it obtained the data. Our approach ends with this step after producing the licence compatibility analysis table.

\noindent\textsf{Running example.} For CIFAR-10, there are large discrepancies between what the dataset license allows and what the data source licenses allow. Table~\ref{tab:MDL} shows the final rights that have changed after the license compatibility analysis in red. \texttt{Tagging} under \texttt{Rights to Data} is restricted by the license of several data sources of CIFAR-10 including Google and Flickr’s licenses and, as such, is not permitted in the verified license (as indicated by the red cross in Table~\ref{tab:MDL}). In summary, we find that even though CIFAR-10's license allows all rights to the dataset (even commercial) as long as the paper is cited, the licenses of its data sources are more restrictive, creating potential risks of license compliance violation if the dataset is used for any commercial purposes including training an AI model, or modifying or distributing the dataset.

\noindent\textbf{Scenario assessment.} Using the rights allowed in the previous step as given by the final license compatibility assessment analysis table, we determine if the given dataset can be used for the required commercial usage scenario (e.g., to embed a model trained on the given dataset in a commercial product). If the rights permit the usage scenario, we convert the obligations to AI software requirements and make it a part of the AI software development life cycle so that it can be continuously traced, documented and tested. 

\noindent\textsf{Running example.} If we want to check if the usage scenario of commercializing the model constructed using CIFAR-10 is allowed, we check if the model commercialization rights are present and what are its associated obligations. From Table~\ref{tab:MDL} we find that model commercialization rights are not present, hence the usage scenario of commercializing the model trained on CIFAR-10 will not be permissible.

\begin{table*}[htbp]
  \centering
  \scriptsize

\caption{License interpretation of CIFAR-10 using the \texttt{enhanced MDL} schema. The fields in blue are our enhancements. The values in red are the changed rights after our license compatibility analysis step.}
~\label{tab:MDL}
\begin{tabular}{|p{1.5cm}|p{0.7cm}|p{0.7cm}|p{0.7cm}|p{0.7cm}|p{0.7cm}|p{0.7cm}|p{0.7cm}|p{0.7cm}|}


\hline
\multirow{8}{*}{\textbf{\makecell[l]{License\\ metadata}}} & \multicolumn{2}{c|}{\textbf{Licensor}} & \multicolumn{2}{c|}{\textbf{\textcolor{blue}{\makecell{License\\ name}}}} & \multicolumn{2}{c|}{\textbf{\makecell{Dataset \\name}}} & \multicolumn{2}{c|}{\textbf{\makecell[l]{\textcolor{blue}{Dataset}\\ \textcolor{blue}{version}}}}\\
\cline{2-9}
& \multicolumn{2}{c|}{Alex Krizhevsky} & \multicolumn{2}{c|}{Custom license} & \multicolumn{2}{c|}{CIFAR-10} & \multicolumn{2}{c|}{N/A}  \\
\cline{2-9}
& \multicolumn{8}{c|}{\textbf{\makecell{Credit/Attribution Notice}}} \\
\cline{2-9}
& \multicolumn{8}{c|}{\makecell{Learning Multiple Layers of Features from Tiny Images, Alex Krizhevsky, 2009. }}\\
\cline{2-9}
& \multicolumn{2}{c|}{\textbf{\textcolor{blue}{\makecell{License\\ validity \\period}}}} &  \multicolumn{2}{c|}{\textbf{\textcolor{blue}{\makecell{Liability\\/Warranty}}}} & \multicolumn{2}{c|}{\textbf{\makecell{Designated\\ third\\ parties}}} & \multicolumn{2}{c|}{\textbf{\makecell{Additional \\conditions}}} \\
\cline{2-9}
& \multicolumn{2}{c|}{N/A} &  \multicolumn{2}{c|}{N/A} & \multicolumn{2}{c|}{\makecell{Only\\ by\\ agreement}} & \multicolumn{2}{c|}{None} \\

\hline

\textbf{\makecell[l]{Data \\(standalone)}} & \multicolumn{2}{c|}{\textbf{Access}} & \multicolumn{2}{c|}{\textbf{Tagging}} & \multicolumn{2}{c|}{\textbf{Distribute}} & \multicolumn{2}{c|}{\textbf{Re-represent}} \\
 \hline
\textbf{Rights} & \multicolumn{2}{c|}{\ding{51}} & \multicolumn{2}{c|}{\ding{51} (\textcolor{red}{\ding{55}})} & \multicolumn{2}{c|}{\ding{51} (\textcolor{red}{\ding{55}})} & \multicolumn{2}{c|}{\ding{51} (\textcolor{red}{\ding{55}})}  \\
\hline
\textbf{\textcolor{blue}{Obligations}}
 & \multicolumn{2}{c|}{\makecell{Cite\\ paper}} & \multicolumn{2}{c|}{\makecell{Cite\\ paper}} & \multicolumn{2}{c|}{\makecell{Cite\\ paper}} & \multicolumn{2}{c|}{\makecell{Cite\\ paper}} \\
\hline
 & & & & & \multicolumn{2}{c|}{\textbf{Commercialization}} &\multicolumn{2}{c|}{}  \\
\cline{6-7}
\textbf{\makecell[l]{Data rights\\ in conjunction\\ with model}} & \textbf{\makecell{Bench-\\mark}} & \textbf{\makecell{Re-\\search}} & \textbf{\makecell{\makecell{Publish}}} & \textbf{\makecell{In-\\ternal\\ Use}} & \textbf{\makecell{Out-\\put}} & \textbf{Model} & \multicolumn{2}{c|}{\textbf{\makecell{\textcolor{blue}{Model} \\\textcolor{blue}{Reverse}\\ \textcolor{blue}{Engineer}}}}\\
\cline{1-9}
\textbf{Rights} & \ding{51} & \ding{51} & \ding{51} & \ding{51} & \ding{51} (\textcolor{red}{\ding{55}}) & \ding{51} (\textcolor{red}{\ding{55}}) & \multicolumn{2}{c|}{\ding{51}} \\ 
\hline
\textbf{\textcolor{blue}{Obligations}} & \makecell{Cite\\ paper} & \makecell{Cite\\ paper} &\makecell{Cite\\ paper} & \makecell{Cite\\ paper} & \makecell{Cite\\ paper} & \makecell{Cite\\ paper} & \multicolumn{2}{c|}{\makecell{Cite\\ paper}} \\ 
\hline
\end{tabular}

\end{table*}

\section{Case Study Details} ~\label{sec:casestudy}
In this section, we explain the details of our case study. Through our case study, we assess the potential license compliance violations that we may incur if we use publicly available datasets to build commercial AI software. 

\subsection{Case Study Setup}
First, we collected a list of popular publicly available datasets from several source like Papers with Code~\cite{PWC} and prior study by~\citet{benjamin2019towards}. Then we chose six common datasets from the list that we obtained based on their popularity and likelihood to be used commercially. The datasets that we study are CIFAR-10~\cite{krizhevsky2009learning}, ImageNet~\cite{russakovsky2015imagenet}, Cityscapes~\cite{cordts2015cityscapes}, FFHQ~\cite{karras2019style}, VGGFace2~\cite{cao2018vggface2} and MS COCO~\cite{lin2014microsoft}. For instance, one of our studied dataset ImageNet is hailed as the dataset that changed machine learning~\cite{Quartz2017}. Similarly, CIFAR-10 and Cityscapes datasets are listed as one of the top 10 most used dataset by Papers with Code~\cite{PWC}. Finally, we include FFHQ and VGGFace2 datasets in our study as several model stores use these datasets to pre-train their AI models~\cite{GluonCVYOLOv3,ModzyModel,TensorFlowdeeplabv3}. These datasets are commonly used for tasks like object detection, image recognition, and semantic segmentation. In addition, all of our studied datasets are image datasets. We choose to focus on image datasets in our study as there are several AI applications and it would be extremely hard to generalize across multiple applications. The most common commercial usage scenarios associated with the datasets are: commercially distributing the datasets (DD), Release product with embedded AI models (RPEAI) trained on these datasets and Commercialize the model output (CAI).
\begin{table}[htbp]
  \centering
  \scriptsize
  \caption{Results of applying our approach on the studied datasets.}
  \label{tab:result}
    \begin{threeparttable}
    \begin{tabular}{llrlrlr}
    \toprule
    \multicolumn{1}{c}{\multirow{3}[6]{*}{\textbf{Dataset}}} & \multicolumn{6}{c}{\textbf{Commercial usage scenario}} \\
\cmidrule{2-7}          & \multicolumn{2}{c}{\textbf{DD}} & \multicolumn{2}{c}{\textbf{RPEAI}} & \multicolumn{2}{c}{\textbf{CAI}} \\
\cmidrule{2-7}          & \textbf{\makecell[l]{Yes\\(obligations)} }  & \multicolumn{1}{l}{\textbf{No}} & \textbf{\makecell[l]{Yes\\(obligations)} }  & \multicolumn{1}{l}{\textbf{No}} & \textbf{\makecell[l]{Yes\\(obligations)} } & \multicolumn{1}{l}{\textbf{No}} \\
    \midrule
    CIFAR-10 &       & \multicolumn{1}{l}{\ding{55}} &       & \multicolumn{1}{l}{\ding{55}} &       & \multicolumn{1}{l}{\ding{55}} \\
    ImageNet &       & \multicolumn{1}{l}{\ding{55}} &       & \multicolumn{1}{l}{\ding{55}} &       & \multicolumn{1}{l}{\ding{55}} \\
    Cityscapes &       & \multicolumn{1}{l}{\ding{55}} &       & \multicolumn{1}{l}{\ding{55}} &       & \multicolumn{1}{l}{\ding{55}} \\
    FFHQ  & \multicolumn{1}{l}{C+D} &       &       & \multicolumn{1}{l}{\ding{55}} &       & \multicolumn{1}{l}{\ding{55}} \\
    VGGFaces2 & \multicolumn{1}{l}{A+E+D} &       &       & \multicolumn{1}{l}{\ding{55}} &       & \multicolumn{1}{l}{\ding{55}} \\
    MS COCO\textsuperscript{*} &  &   \multicolumn{1}{l}{\ding{55}}    &      &   \multicolumn{1}{l}{\ding{55}}    &      & \multicolumn{1}{l}{\ding{55}} \\
    \bottomrule
    \end{tabular}%
    \begin{tablenotes}
    \scriptsize
    \item A -  Provide a link to license CC-BY-NC 4.0; B - Provide a link to the license CC-BY 4.0; C - Provide a link to license CC-By-NC-SA 4.0; D - Remove infringing content as soon as possible when an infringement is detected. E- Indicate changes;
    \item *-Annotations and the images in the dataset are licensed differently, since the dataset is typically used with both the images and annotations, we consider them together. If only annotations are used commercially then the obligations are DD - B+E+D, RPEAI - B, CAI - B
    \end{tablenotes}
  \end{threeparttable}
\end{table}%

We first study the characteristics of the licenses associated with the studied datasets. Then we assess the potential license compliance violations that may occur when using these publicly available datasets by using the approach that we outline in Section~\ref{sec:approach}. 

\subsection{Case study results}

We describe our findings below by first discussing our results regarding the characteristics of the studied datasets in terms of the licenses that they use, and then we discuss the results of our license compliance analysis. 

\smallsection{Characteristics of studied dataset licenses}

\noindent\textbf{[Result 1] 4 out of the 6 studied datasets do not use any standard dataset license (e.g., Creative Commons 4.0).} Among the studied datasets, only FFHQ and VGGFace2 use standard dataset licenses. MS COCO dataset only licenses their annotations under a standard license, they refer to the Flickr terms of use for the license of the images contained in their dataset. Both ImageNet and Cityscapes use a custom license. CIFAR-10 dataset does not explicitly mention any license. The authors of CIFAR-10 only request for a citation from the users of their dataset.

\smallskip\noindent\textbf{[Result 2] The two datasets that use a standard dataset license use licenses from Creative Commons (CC) license family.} FFHQ and VGGFace2 creators license their datasets using CC-NC-SA-4.0 from the CC family of licenses. The MS COCO dataset creators license their dataset's annotations under CC 4.0. 

\smallskip\noindent\textbf{[Result 3] Except Cityscapes dataset (which was created from scratch), all the other studied datasets are created by collecting data from one or more data sources.} In fact, CIFAR-10 and ImageNet datasets have more than one data source from which they collect their data points. Such a result affirms our need for the approach that we outline in Section~\ref{sec:approach} which traces a dataset's lineage and identify its data sources, so the the license compliance can be verified.

\smallsection{License compliance analysis results}

\noindent\textbf{[Result 4] All of the studied publicly available datasets might result in potential license compliance violation if they are used to build commercial AI software.} From Table~\ref{tab:result} we observe that, none of the studied licenses explicitly allow practitioners the right to commercialize an AI model trained on the data or even the output of the trained AI model. Such a result also effectively prevents practitioners from even using pre-trained models trained on these datasets. Publicly available datasets and AI models that are pre-trained on them are widely being used commercially~\cite{peng2021mitigating}. In the case of the MS COCO dataset, only the annotations are licensed under CC 4.0, since users of the dataset typically require the images along with the annotation, using MS COCO dataset under any commercial scenario might result in potential license violation. However, if only the annotations from MS COCO dataset is used to build commercial AI software, they are allowed to do so. However, in such a case practitioners are obligated to provide a link to its license, retain the copyright notice and not use MS COCO dataset's annotations to endorse their product. Failure to do so might once again result in a license compliance violation.

\smallskip\noindent\textbf{[Result 5] 4 out of the 6 studied publicly available datasets might result in potential license compliance violation if the dataset is modified.} More specifically, for CIFAR-10, ImageNet, MS COCO and Cityscapes. Sometimes when a dataset is being used commercially, it is desirable for a dataset to either be modified or enhanced with additional data points or different labels~\cite{shorten2019survey}. For instance, as Cui~et~al.~\cite{cui2015data} show, augmenting the data (i.e., rerepresenting) can be a useful way of improving the performance of an AI model. However, doing so on the above publicly available datasets might result in potential license compliance violation. Even the other datasets, which permit modifications, outline the obligations for practitioners to indicate the exact changes that were made to the dataset. Failure to do so might result in potential license compliance violation.

\smallskip\noindent\textbf{[Result 6] 4 out of the 6 studied publicly available datasets might result in potential license compliance violation if the dataset is distributed along with the commercial AI software.} As we notice from Table~\ref{tab:result}, distribution of CIFAR-10, ImageNet and CityScapes is not allowed. Despite this restriction, many platforms distribute these datasets~\cite{peng2021mitigating}. For instance, DeepAI~\cite{DeepAI} distributes ImageNet and CIFAR-10 even though it is not allowed to do so. Even in the case of the remaining two datasets, where distribution of the dataset is allowed, the obligations outlined by the license state that they must be distributed with the same license of the dataset.

\smallskip\noindent\textbf{[Result 7] In certain cases, it might be necessary to safeguard the model against reverse engineering efforts or adversarial attacks~\cite{rigaki2020survey,song2017machine}.} We note that for the Cityscapes dataset, model reverse engineering is not allowed. If a model trained on Cityscapes data is made available, it must safeguard against attempts that aim to recover the dataset from the trained AI model. Failure to do so might result in a license compliance violation.

\smallskip\noindent\textbf{[Result 8] In addition to rights, all six datasets outline obligations that must be followed. Failure to do so might result in potential license compliance violation.} Table~\ref{tab:result} outline the obligations that one must satisfy when using a given dataset for the outlined commercial usage scenarios. Failure to do so might result in license compliance violation. For instance, ImageNet's license explicitly requires practitioners to indemnify the ImageNet team against any claims arising from use of the dataset. FFHQ, VGGFace2 and MS COCO datasets' annotations require that, if they are distributed or modified, to be presented under the same license. In addition, among the studied datasets we observe that the obligations are standardized (i.e., similar obligations appear across multiple licenses). However, since we studied only 6 datasets, it is challenging to know if our results generalize. 
\hypobox{\textbf{Summary:} Publicly available datasets might not be suitable to build commercial AI software. All of the six studied datasets contain potential risks associated with at least one commercial usage context.}

\section{Feedback from Industry Practitioners}~\label{sec:feedback}

We presented our findings to the practitioners within our AI research teams and gathered feedback. In particular, we present our approach and the analysis results on one of the sample datasets. Then we ask these practitioners to perform license compliance analysis on new datasets of their interests using our approach.

Prior to our proposed approach, whenever the practitioners wished to use a publicly available dataset, they presented that dataset license to the lawyer and enquired if they could use the given dataset commercially. They only seek to interpret the license associated with the dataset without any consideration to the dataset's provenance or lineage. Furthermore, they were simply seeking if they could use the dataset in their AI software or not without distinguishing the type of use (e.g., using the dataset to train the AI model, or distribute the dataset).
After using our proposed approach they realized that for several datasets (e.g., CIFAR-10) they could find the potential license compliance violations only after tracing a dataset's provenance and lineage. 
In addition, they expressed the usefulness of our approach and conveyed that our systematic approach and our case study results can help them demonstrate due diligence. They further expressed a keen interest in pushing our approach for wider adoption in their teams. 
Similarly, all the practitioners expressed the usefulness of the schema that we present as apart of our approach. In particular, they expressed that our enhanced MDL schema provided a great way to standardize our approach and document the rights and obligations associated with using a publicly available dataset. Furthermore, they facilitate better understanding of what commercial use cases can be undertaken.

While majority of the feedback were positive, most of the practitioners expressed concerns over the amount of manual effort involved in our approach. They further stated that the involved manual effort might be prohibitive in some cases to the adoption of our proposed approach. They wished for some automated tools that would help them with the application of our approach to assess license compliance violations. 

\section{Recommendations for AI Engineers}
\label{sec:recommendation}
Based on our case study results and feedback from the industry practitioners, here we provide three recommendations for the AI engineers who use publicly available datasets to create AI software.

\noindent\textbf{Recommendation \#1: Employ caution while using publicly available datasets to build commercial AI software.} 
Just because a dataset is publicly available, it doesn't imply that it could be used to create commercial AI software. Therefore, it is pivotal for the AI engineers to use our approach to verify if the publicly available dataset that they are interested in using can indeed be used in the commercial context that they intend to use.

\noindent\textbf{Recommendation \#2: While assessing the license compliance of datasets, it is essential to use systematic approaches and clear documentations to demonstrate due diligence.}
We recommend the AI engineers to use our proposed approach and the schema to document various aspects (e.g., provenance, lineage and the rights/obligations) of a dataset and its license(s). 

\noindent\textbf{Recommendation \#3: Share the knowledge regarding the risks associated with using a given publicly available dataset commercially.} 
We recommend that AI Engineers to openly share and maintain the use cases where a given dataset can or cannot be used
for the following three reasons: (1) to advocate the importance of this area of work, (2) to minimize the duplicated effort while analyzing the license compliance for the same datasets, and (3) to further validate the analysis via crowd sourcing. 
\section{Threats to Validity}
\label{sec:threats}

In this section, we describe the 
threats that pertain to our study.

\subsection{External validity} 
In our paper we only explore if a given publicly available dataset can be commercially used without violating the license associated with the given dataset. However, we do not explore other factors like privacy and ethical issues that may impact the commercial usage of a given dataset. So even if our process finds that a publicly available dataset maybe commercially used, the user of the the given dataset must verify if other factors might impact its usage. Finally, we only seek to answer if a given publicly available dataset can be used commercially which may not cover other cases such as using the given dataset for internal research or academic purposes. We also do not seek to ascertain if a given dataset can be used under fair use or fair dealing or other similar laws as it typically needs to verified by a court of law. In addition, all of our studied datasets are image datasets. Different datasets like video or text based datasets might pose different challenges. We invite the future research to examine if our proposed approach can be used to assess the potential license compliance risks associated with using those datasets commercially.  

\subsection{Internal validity}
In our study, we only consider the license of the data sources and do not consider the licenses associated with the data points (e.g., individual images) extracted from the data sources. Individual data points could have licenses as well. 
However, each dataset/data source have thousands of data points, making it impossible to extract the licenses of all the individual data points without an automated method. Even if we do manage to extract all their licenses, it is impossible for a lawyer to interpret these licenses and assess their compatibility. Although our proposed approach can demonstrate due diligence, this remains as a threat. 
In addition, 
there can be multiple copies of the same dataset available across different sources with different licenses available. 
To disambiguate and identify the appropriate license(s) associated with a given dataset, our approach specifically extracts a given dataset's provenance and lineage information. 


\subsection{Construct validity} 
In our study, it is possible that the provenance and lineage that we extract for each of the studied datasets is not accurate. For instance, when extracting provenance, we determine the origin of the dataset by checking if the origin has a publication associated with it. However, it is impossible to know for certain if it is indeed a reliable indicator of when (and where) the dataset originated. Similarly, for some of the datasets, such as ImageNet, the exact data sources are unknown. 
In our paper, we make the best effort to infer data sources from the associated papers and documentations so that we can infer their licenses. 

\section{Conclusion}
\label{sec:conclusion}
Publicly available datasets are being widely used to build commercial AI software. One can do so if and only if the license associated with the publicly available dataset provides the right to do so. However, it is not easy to verify the rights and obligations provided in the license associated with the publicly available datasets. Because, at times the license is either unclear or potentially invalid. 

Therefore, in our study, we propose several schema and an approach that allows one to assess and anticipate the potential license compliance violations that may occur when using a publicly available dataset to build AI software. We demonstrate the usefulness of our approach through a case study that we conducted on six commonly publicly available datasets. Through our case study we find the publicly available datasets might not be suitable to build commercial AI software.

\section*{Disclaimer}
Any opinions, findings, and conclusions, or recommendations expressed in this material are those of the author(s) and do not reflect the views of Huawei.
\begin{acks}
We thank Mario Aguilar-Simon from Teledyne Sicentific for his valuable comments regarding the interpretation of MS COCO dataset's license.
\end{acks}

\bibliographystyle{ACM-Reference-Format}
\bibliography{references}


\end{document}